\documentclass[letterpaper]{article} 
\usepackage{aaai23}  
\usepackage{times}  
\usepackage{helvet}  
\usepackage{courier}  
\usepackage[hyphens]{url}  
\usepackage{graphicx} 
\usepackage{natbib}  
\usepackage{caption} 
\usepackage{algorithm}
\usepackage{algorithmic}
\usepackage{amsmath}
\usepackage{amssymb}
\usepackage{subcaption}
\usepackage{makecell}
\usepackage{multirow}
\usepackage{bm}
\usepackage{color,soul}
\usepackage{xcolor}
\definecolor{cerulean}{rgb}{0.0,0.48,0.65}
\definecolor{green}{rgb}{0.01, 0.75, 0.24}
\definecolor{Black}{RGB}{0.0, 0.0, 0.0}

\newcommand{\blue}[1]{\textcolor{Black}{#1}}
\newcommand{\olive}[1]{\textcolor{Black}{#1}}
\newcommand{\cerulean}[1]{\textcolor{Black}{#1}}
\newcommand{\shadow}[1]{}

\def\b{\blue}
\def\o{\olive}
\def\s{\shadow}
\def\c{\cerulean}

\usepackage{fancyhdr}

\usepackage{newfloat}
\usepackage{listings}
\DeclareCaptionStyle{ruled}{labelfont=normalfont,labelsep=colon,strut=off} 
\floatstyle{ruled}
\newfloat{listing}{tb}{lst}{}
\floatname{listing}{Listing}
\usepackage{lipsum}

\title{Self-Supervised Learning for Anomalous Channel Detection in EEG Graphs: Application to Seizure Analysis}
\author{
    Thi Kieu Khanh Ho \textsuperscript{\rm}\thanks{Corresponding Author (Email: thi.k.ho@mcgill.ca).},
    Narges Armanfard
}
\affiliations{
    Department of Electrical and Computer Engineering, McGill University \\
    Mila - Quebec AI Institute, Montreal, QC, Canada \\
}

\usepackage{bibentry}
\usepackage[margin=2cm,top=2.5cm,bottom=2.8cm,headheight=16pt,headsep=0.1in,heightrounded]{geometry}
\usepackage{fancyhdr}
\begin{document}
\pagestyle{fancy}
\fancyhead{} 
\fancyhead[LO, RE]{The paper is accepted for publication at AAAI-23}
\fancyfoot{} 

\maketitle

\begin{abstract}
Electroencephalogram (EEG) signals are effective tools towards seizure analysis where one of the most important challenges is accurate detection of seizure events and brain regions in which seizure happens or initiates. However, all existing machine learning-based algorithms for seizure analysis require access to the labeled seizure data while acquiring labeled data is very labor intensive, expensive, as well as clinicians dependent given the subjective nature of the visual qualitative interpretation of EEG signals. In this paper, we propose to detect seizure channels and clips in a self-supervised manner where no access to the seizure data is needed. The proposed method considers local structural and contextual information embedded in EEG graphs by employing positive and negative sub-graphs. We train our method through minimizing contrastive and generative losses. The employ of local EEG sub-graphs makes the algorithm an appropriate choice when accessing to\s{the} all EEG channels is impossible due to complications such as skull fractures. We conduct an extensive set of experiments on the largest seizure dataset and demonstrate that our proposed framework outperforms the state-of-the-art methods in the EEG-based seizure study. The proposed method is the only study that requires no access to the seizure data in its training phase, yet establishes a new state-of-the-art to the field, and outperforms all related supervised methods.
\end{abstract}

\section{Introduction}\label{introduction}
Epilepsy is one of the most prevalent neurological disorders affecting 65 million people \cite{johnson2019seizures}. Patients with epilepsy suffer from sudden and unforeseen seizures, during which they are vulnerable to injury, suffocation and death \cite{surges2021identifying}. Epileptic seizure detection makes it possible to identify more accurately the epileptogenic zone (EZ), which is the brain area responsible for initiating seizures \cite{sharma2018epileptic}, and the surgical resection of EZ renders epilepsy patients seizure-free. As epilepsy can start at any age, early detection is crucial to prevent further damage during physiological development and to increase life expectancy \cite{beghi2018aging}. 

Scalp electroencephalogram (EEG) has been considered as the most common tool to detect seizures \cite{kuhlmann2018seizure}. EEG measures the voltage changes between electrodes by the ionic currents flowing within the brain neurons and provides spatial-temporal information of the brain. However, detecting seizures in EEGs requires a direct examination by experienced EEG readers that requires a huge amount of time and effort. Moreover, different opinions of experts can cause discrepant diagnostic results \cite{yan2018positive}. Therefore, the development of automated and objective methods for epileptic seizure detection is needed.

Although many studies have proposed deep learning (DL) based models for automated seizure detection \cite{saab2020weak, shoeibi2021epileptic, abdelhameed2021deep, thuwajit2021eegwavenet, khalkhali2021low, rashed2021deep, mahajan2021adopting, saichand2021epileptic, shen2022eeg, gao2022generative}, several challenges still remain unsolved. First, these studies train their proposed model in a supervised approach -- i.e. availability of labeled seizure data during training is required. This does not address the challenge from a clinical perspective as seizure labels obtaining is difficult and labour expensive during the process of manually seizure marking, hence very scarce in real-world applications.

Second, these studies apply deep convolutional neural networks (CNNs) directly to the time-series signals or spectrograms hence ignore the critically important information physical distance-based connectivity and functional-based connectivity between different brain regions. Several studies have recently proposed graph learning techniques to capture the relationships between EEG electrodes (aka EEG nodes) \cite{wang2020weighted, wagh2020eeg, saboksayr2021eeg, tang2021self, saboksayr2021eeg, mathur2022graph}. However, they all fail to take into consideration the \emph{local} patterns (e.g. local sub-graphs and sub-structures) when learning EEG graphs. Such local information could be effective when detecting anomalies in EEG graphs. The effectiveness of such local information has been recently demonstrated in applications such as analyzing social, traffic, citation, and financial transaction networks \cite{liu2021anomaly, liu2021transformer, zheng2021generative}. However, its effectiveness in analyzing EEG signals, that are inherently non-stationary and dynamic, has remained unexplored to date.

Third, as is discussed before, in real-world applications, there is not enough training data from seizure class; while the existence of adequate data from both normal and seizure classes is essential when training supervised algorithms. To handle the imbalance-data issue, some graph-based methods employ graph augmentation \cite{wagh2020eeg, tang2021self}. However, in EEG graphs, not every augmentation technique is effective as some may corrupt the underlying brain region connectivities. Hence, pinpointing appropriate augmentation strategies in EEG graphs that preserve the underlying semantic information is necessary towards proving accurate seizure detection and localization.

Based on the above observations, we propose a novel approach for the detection of abnormal brain regions and EEG channels, called Contrastive and Generative Self-supervised Learning for EEG Graphs (EEG-CGS). Although the effectiveness of contrastive learning methods, as a self-supervised learning technique, has been demonstrated for \o{anomaly detection \cite{li2021cutpaste, hojjati2022self} and graph analysis in general \cite{zeng2021contrastive, xie2022self}}, its ability on analysing the EEG graphs has remained unexplored.

The main contributions of this paper are as follows:

\begin{itemize}
\item We propose a self-supervised method for identifying abnormal brain regions and EEG channels without having access to the abnormal class data during the training phase. To the best of our knowledge, this is the first study for unsupervised identification of abnormal EEG channels and regions.

\item We propose to model brain regions and their connectivities using attributed graphs where each EEG channel is considered as a graph node. Each node is associated with a feature vector constructed from the corresponding EEG signal. Nodes are connected based on four different metrics including nodes Euclidean distance, randomly connection of nodes, node features correlations, and directed transfer function; the first two are meant to capture the geometry of EEG channels and the last two are for capturing connectivity of brain regions.

\item \s{The proposed self-supervised method is realized through employing contrastive and generative learning techniques.}We propose an effective augmentation approach to create the positive and negative pairs\s{required to form the contrastive and generative losses}. Augmentations are based on sub-graphs hence allowing to capture the local structural and contextual information.

\item \c{The proposed self-supervised method is realized through employing contrastive and generative learning techniques.} We define a channel-based anomaly score function which is a linear combination of the contrastive and reconstruction losses.

\item Performance of the proposed abnormal EEG node and region detection is demonstrated on the largest and most comprehensive EEG seizure dataset TUSZ. We show that the proposed EEG-CGS establishes a new state-of-the-art on this dataset. EEG-CGS significantly outperforms all existing (supervised) seizure detection techniques.

\item The proposed technique can be considered as a general approach for the detection and localization in other brain disorders. The sub-graph based nature of EEG-CGS makes it a suitable choice for applications where not all EEG channels are available or reliable during training. E.g. in coma outcome prediction application, the scalp of comatose patients is usually uneven with fractions so that recording of all EEG channels is rarely possible \cite{armanfard2018machine}.
\end{itemize}

\section{Proposed Method} \label{method}

\begin{figure*}[!t]
\centering
\includegraphics[width=0.9\linewidth]{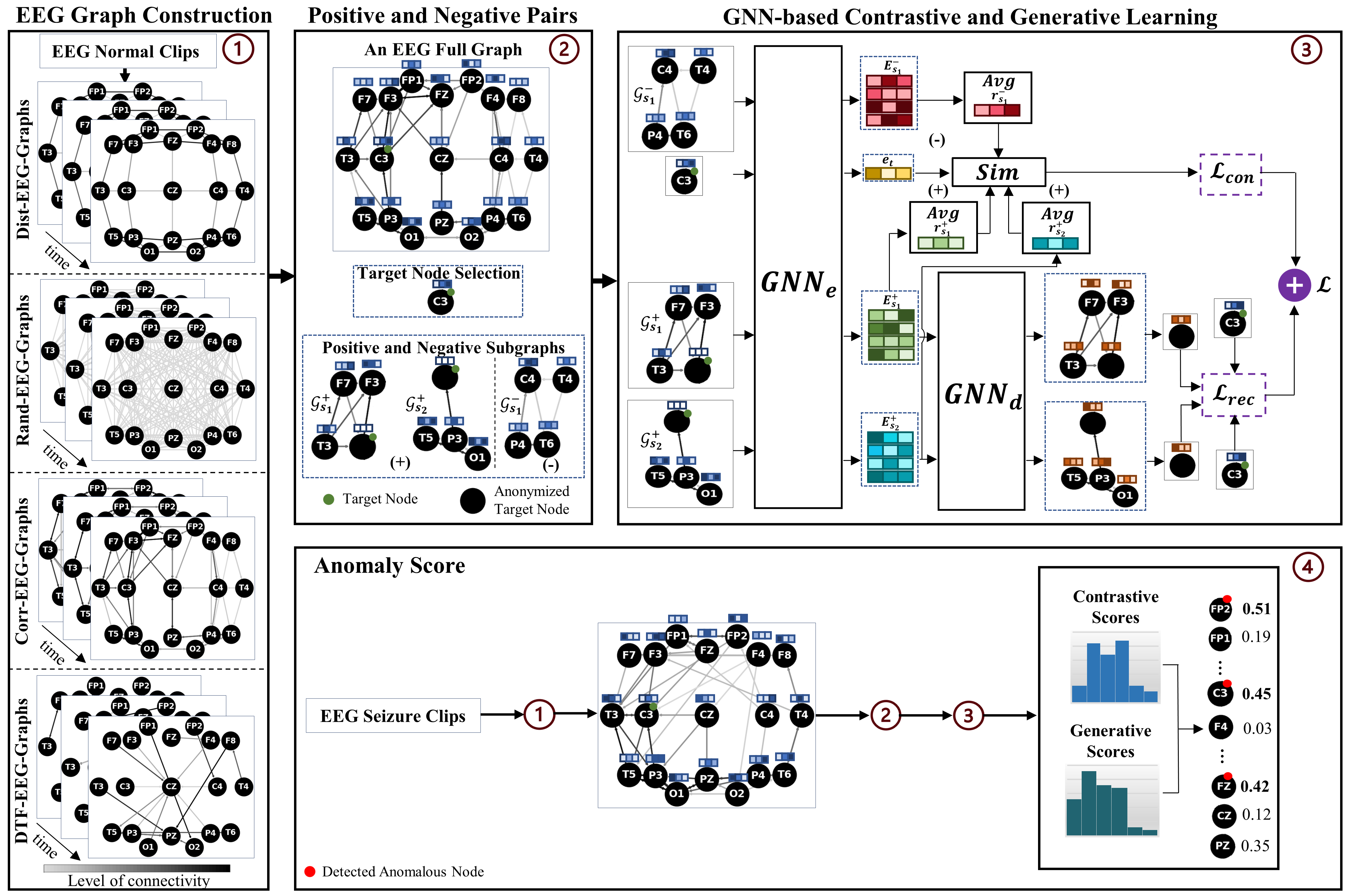}
\caption{The overall framework of EEG-CGS  where $d'=3$, $\alpha=4$ and $\tau_1=0.4$. The positive and negative pairs are shown for a typical target node C3.}
\label{model}
\end{figure*}

The block diagram of the proposed method is shown in Figure \ref{model}.\s{\footnote{The pseudo-code is shown in Appendix.}} The proposed EEG-CGS for node anomaly detection consists of four components namely EEG graph construction, positive and negative pair sampling, contrastive and generative learning based on graph neural networks (GNN), and anomaly score computation. The following sections discuss these components in detail.

\subsection{EEG Graph Construction} \label{graph construction}

An EEG graph is denoted as $\mathcal{G} = (\mathcal{A}, \mathcal{X})$ where $\mathcal{A} = (\mathcal{V}, \mathcal{E})$. $\mathcal{A} \in \mathbb{R}^{n \times n}$ is the adjacency matrix which its element on the $i^{th}$ row and $j^{th}$ column, i.e. $a_{ij}$, is greater than zero if a connectivity exists between EEG electrodes $v_i$ and $v_j$. $\mathcal{V}=\{v_1, ..., v_n\}$ is the set of $n$ electrodes (aka nodes), $\mathcal{E}$ is the set of $m$ edges, $\mathcal{X} \in \mathbb{R}^{n \times d}$ is the feature matrix, $d$ denotes the number of attributes (aka features) representing an EEG signal. The $i^{th}$ row of $\mathcal{X}$, denoted by $x_i$, is the feature vector representing the $i^{th}$ EEG channel. For a given EEG clip, we build four types of EEG graphs:

\begin{itemize}
    \item \emph{Dist-EEG-Graph} endeavors to embed the structure of electrode locations in the graph's adjacency matrix, using Euclidean distance between electrodes. As the electrode locations are fixed in an EEG recording cap, the same adjacency matrix is obtained for all EEG clips. More specifically, $a_{ij}$ of Dist-EEG-Graph is obtained as follows:
    \begin{equation}
    a_{ij} =\left\{\begin{matrix} \exp (-\frac{\left \| v_i - v_j \right \|^2}{\zeta^2}) & \text{if} \,\,\, \left \| v_i - v_j \right \| \leq k\\ 0& \text{if} \,\,\, O.W.\\
    \end{matrix}\right.,
    \end{equation}
    where $\left \| \cdot \right \|$ denotes $\ell_2$-norm and $\zeta$ is a scaling constant. The closer the two electrodes $v_i$ and $v_j$ are, the closer $a_{ij}$ is to 1.  In all our experiments $k$ is set to 0.9 for all EEG clips. Setting $a_{ij}=0$ for far away electrodes imposes sparsity to the graph. 
    
    \item \emph{Rand-EEG-Graph} is built based on the assumption that all electrodes are connected and equally contribute in brain activities. This graph is realized through forming an adjacency matrix as follows: 
    \begin{equation}
    a_{ij} =\left\{\begin{matrix} 0.5 & \text{if} \,\,\, i \neq j\\ 1& \text{if} \,\,\, O.W.\\
    \end{matrix}\right. .
    \end{equation}
    In this way, every edge has the chance of being present in the graph.
    
    \item \emph{Corr-EEG-Graph} is meant to capture the functional connectivity between electrodes. It is encoded in the adjacency matrix elements defined as below:
    \begin{equation}
    a_{ij} =\left\{\begin{matrix} \frac{\text{xcorr}(x_i,x_j)}{\left \| x_i \right \|\left \| x_j \right \|} & \text{if} \,\,\, v_j\in \mathcal{N}(v_i) \\ 0& \text{if} \,\,\, O.W.
    \end{matrix}\right.,
    \end{equation}
    where xcorr($\cdot$,$\cdot$) denotes cross correlation function, $\mathcal{N}(v_i)$ presents the top-3 neighborhood nodes of $v_i$ that have the highest normalized correlation. $\mathcal{N}(v_i)$ is set to the top-3 neighborhood nodes to avoid overly connected graphs. We only keep the top three edges (i.e., $m=3$) for each node to avoid overly connected graphs.
    
    \item \emph{DTF-EEG-Graph} is Directed Transfer Function Graph \cite{blinowska2011review} that is meant to capture the mutual influence between EEG channels, hence it models functional connectivity of the brain regions. The adjacency matrix for this graph is as follows:
    \begin{equation}
    a_{ij} =\left\{\begin{matrix} \frac{\text{xcorr}(x_i,x_j)}{\sqrt{\sum_{m=1 , m \neq i, j}^{n}\|\text{xcorr}(x_i,x_m)\|^2}} & \text{if} \,\,\, v_j\in \mathcal{N}(v_i) \\ 0& \text{if} \,\,\, O.W.\\
    \end{matrix}\right.,
    \end{equation}
    $a_{ij}$ takes a value from 0 to 1, and denotes a ratio between the correlation of the attributes of $v_i$ and $v_j$ to the correlations of the features of all the remaining nodes and $x_i$. As similar to Corr-EEG-Graph, we set $\mathcal{N}(v_i)$ to the top-3 neighbors of $v_i$.
\end{itemize}

\subsection{Positive and Negative Pair Sampling} \label{pos_neg_graph}

For every EEG clip in an input mini-batch $\mathcal{B}$, we first construct the four graphs discussed in the EEG Graph Construction section (see the EEG Graph Construction block in Figure \ref{model}). Then, we create positive and negative sub-graphs for each clip to be used in its following step where the algorithm's networks are trained using reconstruction and contrastive losses. 

We create two positive and one negative sub-graphs for every node in every constructed EEG graph. To this end, we first select an electrode as target node $v_t$ and then select the two positive sub-graphs $\mathcal{G}_{s_1}^+$ and $\mathcal{G}_{s_2}^+$, and one negative sub-graph $\mathcal{G}_{s_1}^-$. Each target node is assigned to its own sub-graphs. To form $\mathcal{G}_{s_1}^+$ and $\mathcal{G}_{s_2}^+$, we leverage a random walk with restart (RWR) technique \cite{tong2006fast} by which sub-graphs are centered at the target node with the fixed size $\alpha$ which controls the radius of the surrounding contexts. In the other words, $\alpha$ determines the number of nodes within the sub-graphs. To impose generalization to the system, we anonymize the target node in the positive sub-graphs by replacing its feature vector with an all-zero vector. To form the negative sub-graph $\mathcal{G}_{s_1}^-$, we first find the farthest electrode from the target node and then apply the RWR to the farthest node with a same $\alpha$ value as of the one used for the positive sub-graphs. In Figure \ref{model}, $\alpha$ is set to 4 indicating 4 nodes are considered in the positive and negative sub-graphs where the farthest node to the target node (i.e. $C3$) is $T4$.

\subsection{Contrastive and Generative Learning} 
As is shown in Figure \ref{model}, we employ GNN autoencoder to realize the unsupervised contrastive and generative learning objectives. The autoencoder consists of two networks: encoder and decoder, respectively denoted as $\text{GNN}_e(\cdot,\cdot)$ and $\text{GNN}_d(\cdot,\cdot)$. Encoder takes a graph as input and embed it to a lower dimensional space while decoder takes the encoder's output as input and endeavors to reconstruct the encoder's input.

Assume a sub-graph $\mathcal{G}_s=(\mathcal{A}_s, \mathcal{X}_s)$, where $\mathcal{A}_s \in \mathbb{R}^{\alpha \times \alpha}$ and $\mathcal{X}_s \in \mathbb{R}^{\alpha \times d}$ respectively denote the sub-graph adjacency matrix and the node attributes. The encoder takes $\mathcal{G}_s$ as input and embed it to a lower dimensional space as below:
\begin{equation}
    E_s = \text{GNN}_e(\mathcal{X}_s, \mathcal{A}_s) \triangleq
 \text{ReLu} (\hat{\mathcal{A}_s}\mathcal{X}_s W_{e}),
    \label{eq: enc}
\end{equation}
where $E_s \in \mathbb{R}^{\alpha \times d'}$, $d'\ll d$, $W_{e}$ denotes the encoder trainable parameters, and ReLu is the Rectified Linear Unit. $\mathcal{\hat{A}}_s = \mathcal{\tilde{D}}_s^{-\frac{1}{2}}\mathcal{\tilde{A}}_s\mathcal{\tilde{D}}_s^{-\frac{1}{2}}$, $\mathcal{\tilde{A}}_s = \mathcal{A}_s + I$ is the adjacency matrix with self-loop, $\mathcal{\tilde{D}}_s$ is the diagonal node degree matrix of the sub-graph where its $i^{th}$ diagonal element is equal to $ \sum_j \tilde{a}_{s_{ij}}$.  $I$ is the identity matrix. For more details, see \cite{zhang2021semi}.

We also embed the target node to the lower dimensional space defined by the encoder. Since a single node $v_i$ has no adjacency matrix, we define its embedding as below:
\begin{equation}
e_i = \text{ReLu}(x_i W_{e}). \label{eq: single-enc}
\end{equation}

In summary, for each constructed graph, we feed three sub-graphs and one target node to the GNN$_e$ to create their embeddings in the lower dimensional space, as is illustrated in the output of GNN$_e$ in Figure \c{\ref{model}} where $d'$ is set to 3.

\subsubsection{Contrastive Learning Module:}
To learn the local topological structure of the graphs, we propose to employ contrastive learning module. In order to create positive and negative \emph{pairs} to be used for the purpose of contrastive learning, we first take an average over all the $\alpha$ rows of $E_s$ to map it to a $d'$ dimensional vector, called $r_s \in \mathbb{R}^{1 \times d'}$.

Since we have formed two positive and one negative sub-graphs for each target node, we define two positive and one negative pairs. The positive pairs are $\mathcal{P}^+_1=(e_t,r_{s_1}^+$) and $\mathcal{P}^+_2=(e_t,r_{s_2}^+$), and the negative pair is $\mathcal{P}^-_1=(e_t,r_{s_1}^-$). Embedding of the target node $v_t$ is denoted by $e_t$, $r_{s_{\{1,2\}}}^+$ indicates the $d'$ dimensional vector obtained by taking an average of the $E_{s_{\{1,2\}}}^+= \text{GNN}_e(\mathcal{G}_{s_{\{1,2\}}}^+)$, and $r_{s_{1}}^-$ indicates the $d'$ dimensional vector obtained by taking average of the $E_{s_{1}}^-= \text{GNN}_e(\mathcal{G}_{s_1}^-)$.

We introduce a trainable scoring matrix $W_s \in \mathbb{R}^{d' \times d'}$ where the similarity of $e_t$ to $r_{s_1}^+$ is obtained as below:
\begin{equation}
\text{Sim}_{t,1}^+=\sigma(e_t W_s r_{s_1}^{+\top}) \label{eq:sim},
\end{equation}
where $\sigma(.)$ is the logistic function that outputs a value between 0 and 1 and $(\cdot)^\top$ denotes the transpose operator. Similarly, $\text{Sim}_{t,2}^+$ and $\text{Sim}_{t,1}^-$ can be obtained by replacing $r_{s_1}^+$ in \eqref{eq:sim} with $r_{s_2}^+$ and $r_{s_1}^-$, respectively. 

Finally, inspired by \cite{zheng2021generative}, we define the contrastive loss function, $\mathcal{L}_{con}$ as below:
\begin{equation}
   \mathcal{L}_{con} = -\frac{1}{2 n |\mathcal{B}|}\sum_{q=1}^{2}\sum_{t=1}^n \Big(  \log(\text{Sim}^{+}_{t,q}) + \log (1-\text{Sim}^{-}_{t,1}) \Big),
\end{equation}
where $|\mathcal{B}|$ indicates the number of EEG clips in the mini-batch $\mathcal{B}$, and $n$ denotes the number of all existing EEG electrodes (e.g., $n$ is 19 in a 10-20 EEG cap recording system). 

\subsubsection{Generative Learning Module:}
This module is to learn the contextual information hidden in the graph through reconstructing the target node anonymized in the positive sub-graphs $\mathcal{G}_1^+$ and $\mathcal{G}_2^+$, using the other node features and edges of the sub-graph. More specifically, we reconstruct the input sub-graph $\mathcal{G}_{s_{\{1,2\}}}^+$ using its embedded version $E_{s_{\{1,2\}}}^+$, i.e. $\hat{\mathcal{G}}_{s_{\{1,2\}}}^+=\text{GNN}_d ( E_{s_{\{1,2\}}}^+ ) $ while minimizing the below reconstruction loss:
\begin{equation}
    \mathcal{L}_{rec} = \frac{1}{2 n |\mathcal{B}|}\sum_{q=1}^2 \sum_{t=1}^n \left \|  \hat{x}_{t,q} - x_t \right \|^2,
\end{equation}
where $\hat{x}_{t,q}$ is the reconstructed vector of the target node $x_t$, taken from the reconstructed graph $\hat{\mathcal{G}}_{s_q}^+=\text{GNN}_d ( E_{s_q}^+ )$. We denote the parameters of the decoder as $W_d$. 

Finally, we define the total loss $\mathcal{L}$ through weighted sum of the reconstruction and contrastive losses, as below:
\begin{equation}
 \mathcal{L} = \lambda \mathcal{L}_{con} + (1 - \lambda) \mathcal{L}_{rec}, \label{eq: loss}
\end{equation}
where $\lambda$ is the balancing  hyperparameter.

\subsection{Anomaly Score}
From the contrastive and generative modules, for a node $v_i \in \mathcal{V}$, we define the anomaly scoring function as below:
\begin{equation}
    \setlength{\abovedisplayskip}{2pt}
    \setlength{\belowdisplayskip}{2pt}
    f(v_i) = \lambda f_{con}(v_i) + (1 - \lambda) f_{rec}(v_i),
\end{equation}
where $f_{con}(v_i) = \frac{1}{2} \sum_{q=1}^{2}\delta(\text{Sim}_{i,1}^{-} - \text{Sim}_{i,q}^{+})$ is the contrastive anomaly score, and $f_{rec}(v_i) = \frac{1}{2}\sum_{q=1}^{2}\delta(||\hat{x}_{i,q} - x_i||^2)$ is the reconstruction anomaly score, and $\delta$ is the MinMaxScaler function \cite{bisong2019introduction} that scales the scores to the range $[0, 1]$. 

The anomaly score for all nodes is calculated. If $v_i$ is an anomaly node from structural point of view, both $\text{Sim}_{i,q}^{+}$, $q \in \{1,2\}$ and $\text{Sim}_{i,1}^{-}$ would be close to 0.5 as there exists a mismatch between $v_i$ and the corresponding node in sub-graphs $\mathcal{G}_{s_q}^+$,$\mathcal{G}_{s_1}^-$ used for training the algorithm networks, because we only use normal clips and nodes in the training phase. Here, the scaling function $\delta$ maps 0 to 1, hence $f_{con}(v_i)$ would be equal to 1. If $v_i$ is an anomaly node from contextual point of view, $f_{rec}(v_i)$ would be close to 1 as the decoder cannot reconstruct an anomaly since the networks are trained on only normal nodes and clips.
If $v_i$ is a normal node, both of $f_{con}(v_i)$ and $f_{rec}(v_i)$ would be close to 0. In other words, an anomaly (normal) node would result into a high (low) anomaly score.

In the inference phase, a node is considered as an anomaly if its corresponding anomaly score is higher than a threshold $\tau_1$. An EEG clip is considered as seizure-free (i.e. normal) if there are less than $\tau_2$ abnormal nodes within its corresponding EEG graph, otherwise it is detected as a seizure clip. The inference phase for a seizure clip is illustrated in Figure \ref{model} at block (4).

\section{Experiments}
We use the public benchmark dataset, Temple University Hospital EEG Seizure Corpus (TUSZ) v1.5.2 \cite{shah2018temple}, which contains the largest number of seizure categories and patient samples. TUSZ is recorded over years, and recorded by different equipment generations from subjects of all ages. Thus, the variance is much more than other available EEG datasets in the seizure study, hence it is the most challenging dataset for seizure detection. The number of existing channels is 19 recorded in the standard EEG 10-20 system, as is shown in Figure \ref{model}. Table \ref{data} summarizes the TUSZ dataset used in our experiments. As is shown in the table, in the training phase, we use the same number of normal clips as is used in other supervised methods, and do not use any seizure clips. In the test phase, the same number of test clips, including seizure and normal clips, is used in our comparison supervised methods and our method. To evaluate the model's ability in seizure location detection, we use the available annotations on focal and generalized seizure types from 23 distinct patients. Note that compared with other types of seizures, focal and generalized types are most present in epilepsy patients. 

\begin{table}[t]
\centering
\scalebox{0.9}{
\begin{tabular}{c | c | c | c } 
     \hline
     \textbf{Data} & \makecell{\textbf{Patients} \\ (\% SZ)} & \makecell{\textbf{EEG Files} \\ (\% SZ)} & \makecell{\textbf{EEG Clips} \\ (\% SZ)} \\ 
     \hline\hline
     Train$_\text{Supervised}$ & 591 (34.0\%) & 4,599 (18.9\%) & 38,613 (9.3\%)\\
     \hline
     Train$_\text{EEG-CGS}$ & 493 (0\%) & 4,028 (0\%)& 35,019 (0\%) \\ 
     \hline
     Test & 45 (77.8\%) & 900 (25.6\%) & 8,848 (14.7\%) \\
     \hline
\end{tabular}}
\caption{{Train and test sets of TUSZ used in the supervised method and EEG-CGS}. The percentages of the seizure data (SZ) is indicated in parenthesis.}
\label{data}
\end{table}

Performance of the proposed method for anomaly detection is quantified using five criteria including Receiver Operating Characteristics - Area Under the Curve (ROC-AUC), Precision (Pre.), F1 score (F1), Sensitivity (Sen.) and Specificity (Spec.). 

We compare our proposed EEG-CGS with two streams of DL-based methods \footnote{To provide a fair comparison, we do not include algorithms that require training their model on an extra EEG dataset (using transfer learning), in addition to the given TUSZ data.}: (1) DL models in the EEG time-series and/or spectrograms domain, including EEGNet \cite{lawhern2018eegnet}, EEG-TL \cite{khalkhali2021low}, Dense-CNN, LSTM and CNN-LSTM \cite{tang2021self}; and (2) DL models in the EEG graph domain \cite{tang2021self}. Unlike our method, these DL models make use of the seizure data and their corresponding labels in the training phase. For a fair comparison, we test the methods on the same test data, whilst our method is trained on a much smaller number of training sample because we do not include any seizure data in the training phase.

We introduce four variations of our EEG-CGS method based on the type of constructed graph used as the input. In the following, we refer to these variations as EEG$_{d}$-CGS, EEG$_{r}$-CGS, EEG$_{c}$-CGS, EEG$_{f}$-CGS and EEG$_{t}$-CGS, respectively corresponding to the cases where the employed input graph is Dist-EEG-Graph, Rand-EEG-Graph, Corr-EEG-Graph, DTF-EEG-Graph and the case where all the four graph types are concatenated and fed to the system as the input representing the given input EEG clip. In our experiments, following \cite{tang2021self}, we simply use the Fast Fourier Transform coefficients as attributes of EEG channels \s{\footnote{\orange{See Appendix for more details on data pre-processing.}}}, although adding other attributes such as Wavelet coefficients might also be useful, which is out of the scope of this paper. 

The hyperparameters $d, \, k, \, m, \, d^{'}, \, \alpha, \, \lambda, \, \tau_1, \, \tau_2$ of the proposed method are respectively set to 6000, 0.9, 3, 256, 4, 0.6, 0.4, 1. \s{\footnote{\orange{Hyperparameters of EEG-CGS are presented in Appendix.}}} To have a fair comparison, all hyperparameters of the proposed method are fixed over all experiments. Regarding the EEG$_{t}$-CGS case, we also set the hyperparameters to be the same as the values mentioned above; where at the inference phase, we take an average of anomaly scores computed for all nodes of the four graph types -- a node is predicted as an anomaly if the average score is greater than $\tau_1$. Based on the average anomaly scores for all nodes, an EEG clip is predicted as seizure if the number of predicted anomalous nodes (over all clips) is greater than $\tau_2$. Hyperparameters of our comparison methods are set to the values suggested by their authors.

The performance of our proposed method is demonstrated for anomalous channel detection, seizure clip detection and seizure channel detection.

\begin{table}[t]
\centering
\scalebox{0.9}{
\begin{tabular}{c | c | c | c | c} 
     \hline
     \textbf{Graph Type} & \textbf{Pre} & \textbf{F1} & \textbf{Sen} & \textbf{Spec}\\ 
     \hline\hline
     EEG$_{d}$-CGS & 0.846 & 0.771 & 0.688 & 0.945\\ 
     \hline
     EEG$_{r}$-CGS & 0.877 & 0.881 & 0.887 & 0.979 \\ 
     \hline
     EEG$_{c}$-CGS & 0.861 & 0.843 & 0.825 & 0.953\\ 
     \hline
     EEG$_{f}$-CGS & 0.844 & 0.832 & 0.822 & 0.967 \\ 
     \hline
     EEG$_{t}$-CGS & $\bm{0.901}$ & $\bm{0.883}$ & $\bm{0.920}$ & $\bm{0.989}$ \\ 
     \hline
\end{tabular}}
\caption{Synthetic anomalous node detection performance. The best performance for each metric is shown in bold.}
\label{node_result}
\end{table}

\subsection{Detection of Synthetic Anomalous Channels}

This section is to verify the performance of the proposed method for reliable anomalous channel detection. To this end, we create a synthetic test set using the normal clips used in the training phase. More specifically, we first average every 35 normal clips (with no overlap). We then, with probability of 3\%, inject an anomaly node to the average clip. A node of the selected average clip will be corrupted with a probability of 0.03\%, and no more than one node is corrupted per average clip. A node is corrupted both contextually and structurally. To this end, we perform two types of corruptions to make $v_i$ abnormal: (1) we connect $v_i$ to all other nodes in the average clip -- i.e. we set $a_{ij}=1$ for $j=1,\ldots,n$; (2) we corrupt the attribute vector of $v_i$ by replacing its feature vector with the feature vector of the node in the average clip that has the largest Euclidean distance to $v_i$. We then feed the average clips (that some have injected anomalies) to the EEG-CGS networks trained on the pure normal clips and let the trained system compute the anomaly score for all channels. If the anomaly score is greater than the pre-defined threshold $\tau_1$, then the corresponding channels are marked as anomalies.

\cite{liu2021anomaly} demonstrated that, for node anomaly detection in the test phase, it would be more effective to create multiple contrastive pairs to capture more neighbors of the target node. (Note that in the training phase, we created only two contrastive pairs.) As such, following \cite{liu2021anomaly}, we compute the anomaly score for 80 randomly selected sub-graphs that include both close and far nodes around the target node. The final anomaly score of a target node is then the average of the 80 calculated ones. This strategy makes the final anomaly score more reliable. Our experiments show high detection performance can be obtained for a wide range of contrastive pair numbers.

Node anomaly detection results for the four variations of our EEG-CGS method are reported in Table \ref{node_result}. The results demonstrate the effectiveness of our unsupervised method for anomalous EEG channel detection. Note that this is the only existing study for anomalous EEG channel detection with access to no abnormal data. The reported results, confirm the effectiveness of the reconstructed graphs and that including all graphs for detection results in a more consistent performance across all metrics.

\begin{table}[t]
\centering
\scalebox{0.9}{
\begin{tabular}{c | c | c | c | c}
\hline
\textbf{Approach} & \textbf{Method} & \textbf{F1} & \textbf{Sen} & \textbf{Spec}\\ 
    \hline\hline
    \multirow{7}{*}{\textbf{Supervised}} 
         & \multicolumn{1}{c|}{EEGNet} & \multicolumn{1}{c|}{0.474} & \multicolumn{1}{c|}{0.299} & \multicolumn{1}{c}{0.902}\\\cline{2-5}
         & \multicolumn{1}{c|}{EEG-TL} & \multicolumn{1}{c|}{0.420} & \multicolumn{1}{c|}{NA} & \multicolumn{1}{c}{NA}\\\cline{2-5}
         & \multicolumn{1}{c|}{Dense-CNN} & \multicolumn{1}{c|}{0.404} & \multicolumn{1}{c|}{0.451} & \multicolumn{1}{c}{0.869}\\\cline{2-5}
         & \multicolumn{1}{c|}{LSTM} & \multicolumn{1}{c|}{0.365} & \multicolumn{1}{c|}{0.463} & \multicolumn{1}{c}{0.814}\\\cline{2-5}
         & \multicolumn{1}{c|}{CNN-LSTM} & \multicolumn{1}{c|}{0.330} & \multicolumn{1}{c|}{0.363} & \multicolumn{1}{c}{0.857}\\\cline{2-5}
         & \multicolumn{1}{c|}{\textit{Dist}-DCRNN} & \multicolumn{1}{c|}{0.341} & \multicolumn{1}{c|}{0.326} & \multicolumn{1}{c}{0.932}\\\cline{2-5}
         & \multicolumn{1}{c|}{\textit{Corr}-DCRNN} & \multicolumn{1}{c|}{0.448} & \multicolumn{1}{c|}{0.457} & \multicolumn{1}{c}{0.900}\\\cline{2-5}
         \hline \hline
    \multirow{5}{*}{\makecell{\textbf{Self-}\\\textbf{Supervised}\\\textbf{(ours)}}}
         & \multicolumn{1}{c|}{EEG$_{d}$-CGS} & \multicolumn{1}{c|}{0.487} & \multicolumn{1}{c|}{0.481} & \multicolumn{1}{c}{0.932}\\\cline{2-5}
         & \multicolumn{1}{c|}{EEG$_{r}$-CGS} & \multicolumn{1}{c|}{0.496} & \multicolumn{1}{c|}{0.465} & \multicolumn{1}{c}{0.942}\\\cline{2-5}
         & \multicolumn{1}{c|}{EEG$_{c}$-CGS} & \multicolumn{1}{c|}{$0.521^{\dagger}$} & \multicolumn{1}{c|}{$0.497^{\dagger}$} & \multicolumn{1}{c}{0.942}\\\cline{2-5}
         & \multicolumn{1}{c|}{EEG$_{f}$-CGS} & \multicolumn{1}{c|}{0.516} & \multicolumn{1}{c|}{0.474} & \multicolumn{1}{c}{$0.952^{\dagger}$}\\\cline{2-5}
         & \multicolumn{1}{c|}{EEG$_{t}$-CGS} & \multicolumn{1}{c|}{$\bm{0.534}$} & \multicolumn{1}{c|}{$\bm{0.501}$} & \multicolumn{1}{c}{$\bm{0.974}$}\\\cline{2-5}
         \hline
\end{tabular}}
\caption{Seizure clips detection. The best and the second-best metrics are denoted in bold and $^\dagger$, respectively.}
\label{sz_result}
\end{table}

\begin{table}[t]
\centering
\scalebox{0.9}{
\begin{tabular}{c | c | c | c | c} 
     \hline
     \textbf{Graph Type} & \textbf{Pre} & \textbf{F1} & \textbf{Sen} & \textbf{Spec}\\ 
     \hline\hline
     EEG$_{d}$-CGS & 0.668 & 0.474 & 0.367 & 0.656 \\ 
     \hline
     EEG$_{r}$-CGS & 0.631 & 0.334 & 0.327 & 0.748  \\ 
     \hline
     EEG$_{c}$-CGS & 0.667 & 0.433 & 0.424 & 0.745 \\ 
     \hline
     EEG$_{f}$-CGS & 0.659 & 0.394 & 0.378 & 0.735 \\ 
     \hline
     EEG$_{t}$-CGS & $\bm{0.704}$ & $\bm{0.551}$ & $\bm{0.432}$ & $\bm{0.775}$ \\ 
     \hline
\end{tabular}}
\caption{Seizure channels detection performance. The best metric is in bold.}
\label{sz_location_result}
\end{table}

\begin{figure}[!t]
\centering
\includegraphics[width=0.99\linewidth]{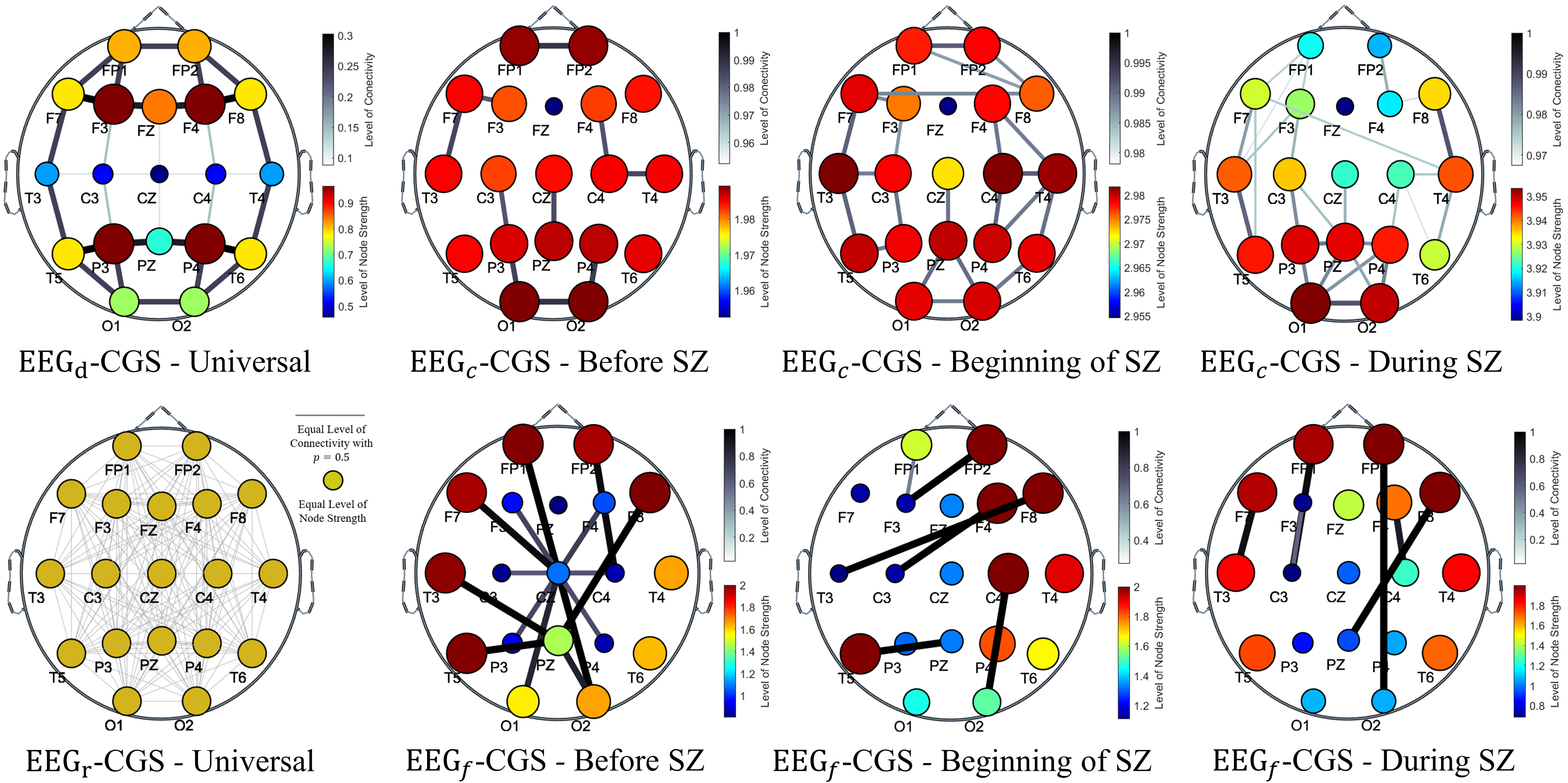}
\caption{Visualization of EEG graphs before, begining and during a typical seizure (SZ) incident. Darker colors indicate higher values.}
\label{eeg_graphs}
\end{figure}

\subsection{Detection of Seizure Clips and Channels}

\begin{figure*}[!t]
\centering
\begin{subfigure}{.49\textwidth}
  \centering
  \includegraphics[width=0.99\linewidth]{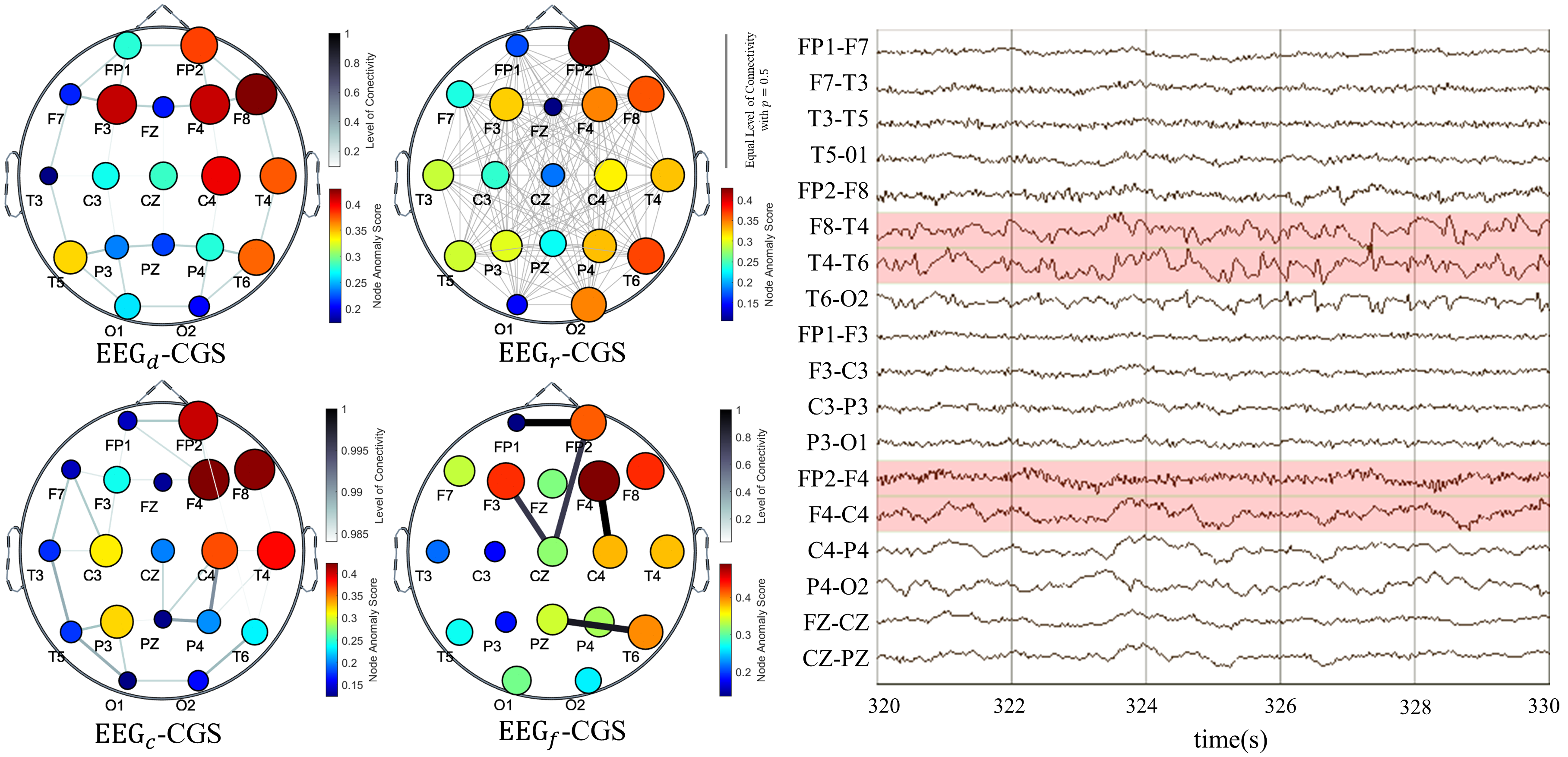}
  \caption{Focal seizure channels detection versus Ground-truth.}
  \label{fig:focal_sz}
\end{subfigure}
\begin{subfigure}{.49\textwidth}
  \centering
  \includegraphics[width=0.99\linewidth]{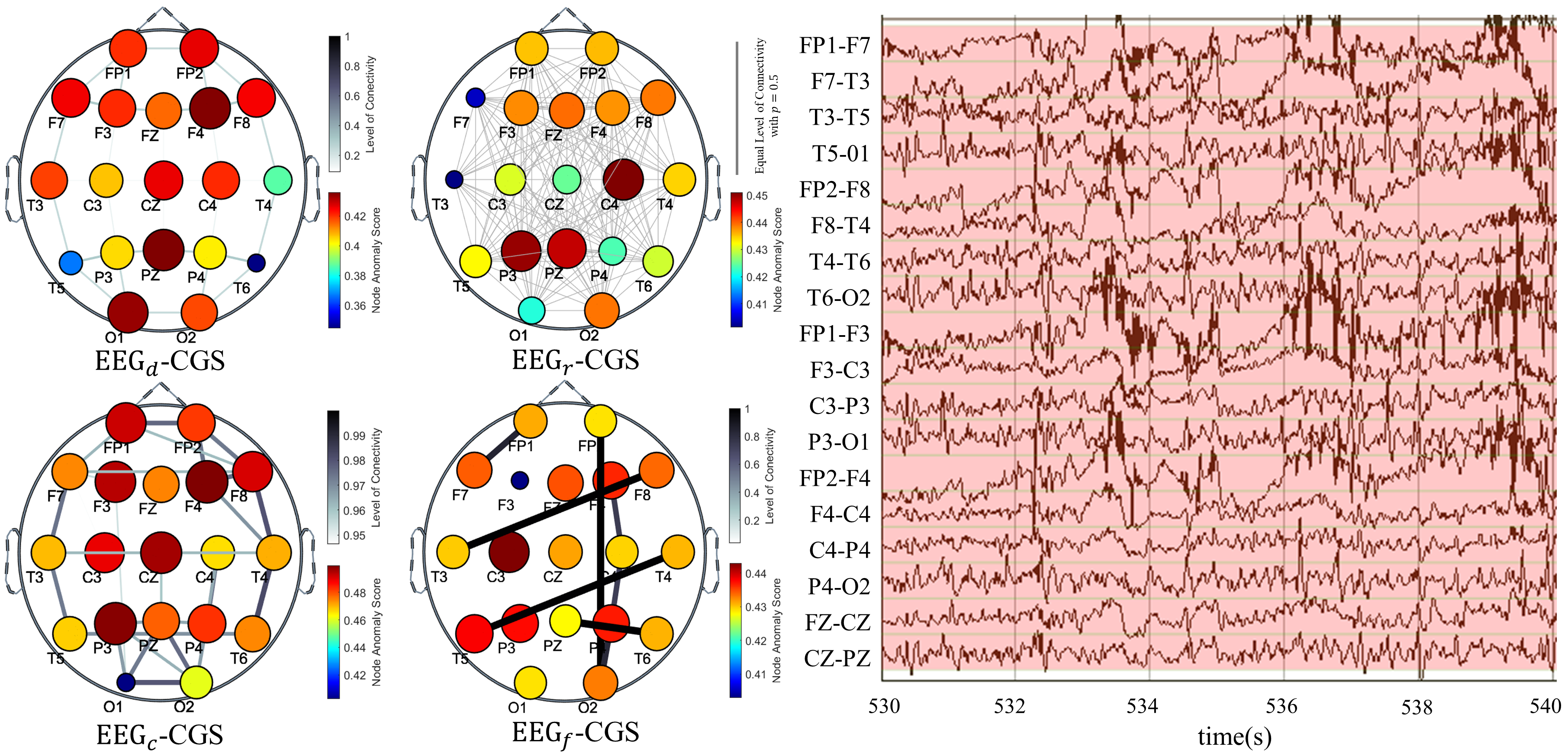}
  \caption{Generalized seizure channels detection versus Ground-truth.}
  \label{fig:gen_sz}
\end{subfigure}
\caption{Visualization of seizure hannels detection for (a) Focal- and (b) Generalized seizures. Ground-truth channels are shaded in pink. Each EEG graph shows the level of the node connectivity and node anomaly score. A darker color indicates higher node connectivity and anomaly score.}
\label{fig:sz_graphs}
\end{figure*}

\setlength{\belowcaptionskip}{-3pt}
\begin{figure*}[!t]
\centering
\begin{subfigure}{.3\textwidth}
  \centering
  \includegraphics[width=0.8\linewidth]{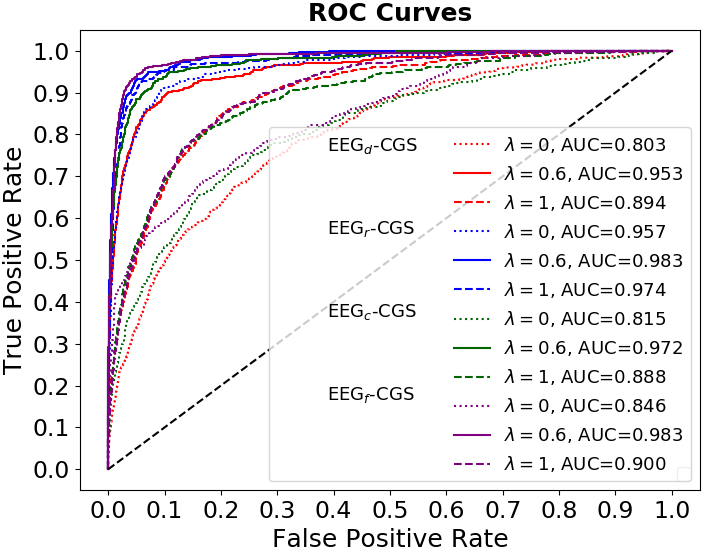}
  \caption{Balancing Factor $\lambda$}
  \label{fig:auc_full}
\end{subfigure}
\begin{subfigure}{.3\textwidth}
  \centering
  \includegraphics[width=.8\linewidth]{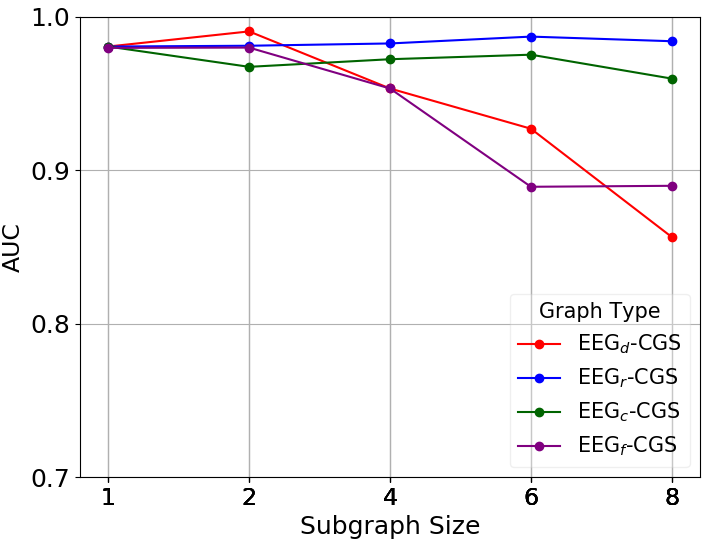}
  \caption{Dimension of Each Subgraph}
  \label{fig:auc_subgraph}
\end{subfigure}
\begin{subfigure}{.3\textwidth}
  \centering
  \includegraphics[width=0.8\linewidth]{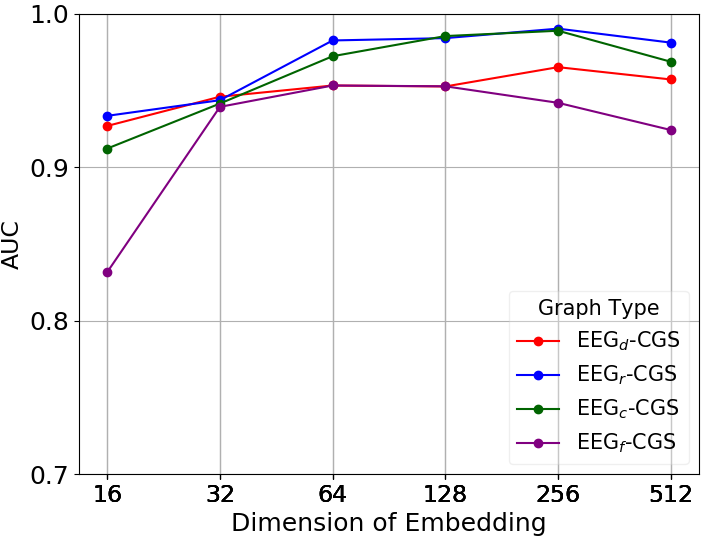}
  \caption{Dimension of Embedding}
  \label{fig:auc_embedding}
\end{subfigure}
\caption{Hyper-parameter sensitivity for synthetic anomalous node detection.}
\label{fig:auc}
\end{figure*}

The performance of the proposed method along with our comparison methods, for seizure clip detection, are shown in Table \ref{sz_result}. Note that the proposed method is a self-supervised 1-class classification technique while all the comparison methods are supervised 2-class classification methods. The compared methods are the current state-of-the-art in the seizure detection research field. Our method significantly outperforms all the comparison methods in all metrics and set a new SOTA in the field though it is trained with only normal data compared to the other methods that are trained with much more data including both normal and abnormal.

To show the effectiveness of our proposed method for anomalous channel detection, in a seizure test clip, we mark the channel with an anomaly score greater than $\tau_1$ as an anomaly and compare them with the ground-truth of seizure channels provided in the TUSZ dataset. The performance is reported in Table \ref{sz_location_result}. As it can be seen, the proposed technique can effectively identify the anomalous node in a real-world dataset, as well.

Figure \ref{fig:sz_graphs} visualizes the anomaly score of the EEG channels for a typical focal seizure clip and a typical generalized seizure clip.  Subfigure (a) shows that, as is expected, the channels with high anomaly scores (e.g., FP2 to C4 and F8 to T6) are focused on specific brain regions that also match the ground-truth. For the generalized seizure case shown in subfigure (b), as is expected, the EEG-CGS method correctly tells us that anomalous channels are spread across the brain, not focused on a specific region as opposed to the focal seizure case shown in (a).

\subsection{Visualization of Graph's Nodes Connectivity}

This section is to visualize the EEG graph level of nodes connectivity (i.e. adjacency matrix) changes when a seizure happens. At each graph, the intensity of an edge between nodes $v_i$ and $v_j$ depends on the value of $a_{ij}$ in the graph's adjacency matrix. This figure also visualizes the level of node strength, where the level of strength at node $v_i$ is defined as $\Sigma_{j=1, j \neq i}^{n} a_{ij}$, i.e. the summation of all connectivities per node -- if a node is connected to channels with high adjacency values, a high level of strength is assigned to the node.

As is discussed in the EEG Graph Construction section, the adjacency matrix of Dist-EEG and Rand-EEG graphs do not change when a seizure happens since they are defined based on the Euclidean distance of node locations and independent from the node attributes. However, the connectivities (i.e. adjacency matrices) vary in the Corr-EEG and DTF-EEG graphs when a seizure happens. These changes are noticeable when one compares the node connectivities and level of strengths before, at the onset, and during a seizure incident. As is shown in Figure \ref{eeg_graphs}, before a seizure incident (i.e. in normal brain activity), in the Corr-EEG-Graph, all nodes have almost similar level of strengths and every node is connected to almost similar number of other nodes. This uniform structure changes when the seizure starts and deepen during a seizure episode. The changes happen to the DTF-EEG graph are also visualized in the figure. As it can be seen, this graph also carries important information and its variations help in distinguishing a seizure activity from normal. This figure demonstrates the effectiveness of the proposed constructed graphs in identifying seizure channels and clips in the self-supervised way.

\subsection{Hyperparameters Sensitivity}
In this section, we investigate the effect of hyperparameters $\lambda$, $\alpha$, and $d'$ on the performance of the proposed method. \s{\footnote{\b{More experiments on the choice of hyperparameters are presented in Appendix.}}}

In  Figure \ref{fig:auc_full}, we explore the importance of the contrastive and generative components through changing the hyperparameter $\lambda$ in the loss function $\mathcal{L}$ defined in \eqref{eq: loss}, where $\lambda \in \{ 0, 0.6, 1\}$. $\lambda=0 \, (1)$ is equivalent to removing the contrastive (generative) component of the proposed method, while setting $\lambda$ to a moderate value such as 0.6 allows the method to include both components when training its networks. We explored the performance of EEG$_{\{d,r,c,f\}}$-CGS for these three values. The results show that for all method variations, $\lambda=0.6$ provides a better AUC compared to the extreme cases $\lambda=0$ and $\lambda=1$.

In  Figure \ref{fig:auc_subgraph}, we explore the impact of subgraph size, i.e. $\alpha$\s{, on EEG-CGS} by selecting $\alpha$ from the set $\{ 1, 2, 4, 6, 8 \}$. We observe that more reliable performance (with high AUC value) can be obtained for $\alpha=4$, i.e. 4 nodes are included in a sub-graph.

To investigate the effect of the embedding space dimension $d'$ on the proposed method, we run our method variations using different $d'$ values from set $\{16, 32, 64, 128, 256, 512\}$. The results shown in Figure \ref{fig:auc_embedding} demonstrate that the method is not too sensitive to this hyperparameter and a reliable and accurate performance can be obtained for a wide range of $d'$.

\section{Conclusion}

This paper is the first attempt on providing a self-supervised method for anomalous EEG channel detection. The proposed EEG-CGS method, consisting of contrastive and generative components, is trained only on normal EEG data with no-access to the abnormal samples. We propose to construct EEG attributed sub-graphs to capture the local structural and contextual information. The performance of the proposed method is demonstrated on both synthetic and real-world anomalies. The proposed method sets a new state-of-the-art for seizure detection on the largest and most difficult publicly available seizure dataset. EEG-CGS can be considered as an anomalous channel detection technique that can be employed for the detection of other brain disorders. The employ of sub-graphs rather than the whole EEG graph also makes the algorithm appropriate in applications, such as coma outcome prognosis, where access to all EEG channels is not possible, e.g. due to skull fractures.

\section{Acknowledgments}
\c{The authors would like to acknowledge the financial support of the Natural Sciences, Engineering Research Council of Canada (NSERC), Fonds de recherche du Quebec, and the Department of Electrical and Computer Engineering at McGill University. The authors also wish to thank Calcul Quebec and Compute Canada for providing the necessary computational resources to conduct our experiments.}

\section{Ethics Statement}
\c{The Temple University Hospital EEG Seizure Corpus is publicly available with full Institutional Review Board (IRB) approval \cite{shah2018temple} at \url{https://isip.piconepress.com/projects/tuh_eeg/html/downloads.shtml}. The authors have no conflicts of interest to declare. No harmful insights are observed by our proposed method described in this paper. Source code is publicly available at  \url{https://github.com/Armanfard-Lab/EEG-CGS}.}

\bibliography{aaai23}

\begin{thebibliography}{35}
\providecommand{\natexlab}[1]{#1}

\bibitem[{Abdelhameed and Bayoumi(2021)}]{abdelhameed2021deep}
Abdelhameed, A.; and Bayoumi, M. 2021.
\newblock A deep learning approach for automatic seizure detection in children
  with epilepsy.
\newblock \emph{Frontiers in Computational Neuroscience}, 15: 650050.

\bibitem[{Armanfard et~al.(2018)Armanfard, Komeili, Reilly, and
  Connolly}]{armanfard2018machine}
Armanfard, N.; Komeili, M.; Reilly, J.~P.; and Connolly, J.~F. 2018.
\newblock A machine learning framework for automatic and continuous MMN
  detection with preliminary results for coma outcome prediction.
\newblock \emph{IEEE journal of biomedical and health informatics}, 23(4):
  1794--1804.

\bibitem[{Beghi and Giussani(2018)}]{beghi2018aging}
Beghi, E.; and Giussani, G. 2018.
\newblock Aging and the epidemiology of epilepsy.
\newblock \emph{Neuroepidemiology}, 51(3-4): 216--223.

\bibitem[{Bisong(2019)}]{bisong2019introduction}
Bisong, E. 2019.
\newblock Introduction to Scikit-learn.
\newblock In \emph{Building machine learning and deep learning models on Google
  cloud platform}, 215--229. Springer.

\bibitem[{Blinowska(2011)}]{blinowska2011review}
Blinowska, K.~J. 2011.
\newblock Review of the methods of determination of directed connectivity from
  multichannel data.
\newblock \emph{Medical \& biological engineering \& computing}, 49(5):
  521--529.

\bibitem[{Gao et~al.(2022)Gao, Zhou, Yang, Chi, and Yuan}]{gao2022generative}
Gao, B.; Zhou, J.; Yang, Y.; Chi, J.; and Yuan, Q. 2022.
\newblock Generative adversarial network and convolutional neural network-based
  EEG imbalanced classification model for seizure detection.
\newblock \emph{Biocybernetics and Biomedical Engineering}, 42(1): 1--15.

\bibitem[{Hojjati, Ho, and Armanfard(2022)}]{hojjati2022self}
Hojjati, H.; Ho, T. K.~K.; and Armanfard, N. 2022.
\newblock Self-Supervised Anomaly Detection: A Survey and Outlook.
\newblock \emph{arXiv preprint arXiv:2205.05173}.

\bibitem[{Johnson(2019)}]{johnson2019seizures}
Johnson, E.~L. 2019.
\newblock Seizures and epilepsy.
\newblock \emph{Medical Clinics}, 103(2): 309--324.

\bibitem[{Khalkhali et~al.(2021)Khalkhali, Shawki, Shah, Golmohammadi, Obeid,
  and Picone}]{khalkhali2021low}
Khalkhali, V.; Shawki, N.; Shah, V.; Golmohammadi, M.; Obeid, I.; and Picone,
  J. 2021.
\newblock Low Latency Real-Time Seizure Detection Using Transfer Deep Learning.
\newblock In \emph{2021 IEEE Signal Processing in Medicine and Biology
  Symposium (SPMB)}, 1--7. IEEE.

\bibitem[{Kuhlmann et~al.(2018)Kuhlmann, Lehnertz, Richardson, Schelter, and
  Zaveri}]{kuhlmann2018seizure}
Kuhlmann, L.; Lehnertz, K.; Richardson, M.~P.; Schelter, B.; and Zaveri, H.~P.
  2018.
\newblock Seizure prediction—ready for a new era.
\newblock \emph{Nature Reviews Neurology}, 14(10): 618--630.

\bibitem[{Lawhern et~al.(2018)Lawhern, Solon, Waytowich, Gordon, Hung, and
  Lance}]{lawhern2018eegnet}
Lawhern, V.~J.; Solon, A.~J.; Waytowich, N.~R.; Gordon, S.~M.; Hung, C.~P.; and
  Lance, B.~J. 2018.
\newblock EEGNet: a compact convolutional neural network for EEG-based
  brain--computer interfaces.
\newblock \emph{Journal of neural engineering}, 15(5): 056013.

\bibitem[{Li et~al.(2021)Li, Sohn, Yoon, and Pfister}]{li2021cutpaste}
Li, C.-L.; Sohn, K.; Yoon, J.; and Pfister, T. 2021.
\newblock Cutpaste: Self-supervised learning for anomaly detection and
  localization.
\newblock In \emph{Proceedings of the IEEE/CVF Conference on Computer Vision
  and Pattern Recognition}, 9664--9674.

\bibitem[{Liu et~al.(2021{\natexlab{a}})Liu, Li, Pan, Gong, Zhou, and
  Karypis}]{liu2021anomaly}
Liu, Y.; Li, Z.; Pan, S.; Gong, C.; Zhou, C.; and Karypis, G.
  2021{\natexlab{a}}.
\newblock Anomaly detection on attributed networks via contrastive
  self-supervised learning.
\newblock \emph{IEEE transactions on neural networks and learning systems},
  33(6): 2378--2392.

\bibitem[{Liu et~al.(2021{\natexlab{b}})Liu, Pan, Wang, Xiong, Wang, Chen, and
  Lee}]{liu2021transformer}
Liu, Y.; Pan, S.; Wang, Y.~G.; Xiong, F.; Wang, L.; Chen, Q.; and Lee, V.~C.
  2021{\natexlab{b}}.
\newblock Anomaly detection in dynamic graphs via transformer.
\newblock \emph{IEEE Transactions on Knowledge and Data Engineering}.

\bibitem[{Mahajan, Somaraj, and Sameer(2021)}]{mahajan2021adopting}
Mahajan, A.; Somaraj, K.; and Sameer, M. 2021.
\newblock Adopting artificial intelligence powered ConvNet to detect epileptic
  seizures.
\newblock In \emph{2020 IEEE-EMBS Conference on Biomedical Engineering and
  Sciences (IECBES)}, 427--432. IEEE.

\bibitem[{Mathur and Chakka(2022)}]{mathur2022graph}
Mathur, P.; and Chakka, V.~K. 2022.
\newblock Graph Signal Processing Based Cross-Subject Mental Task
  Classification Using Multi-Channel EEG Signals.
\newblock \emph{IEEE Sensors Journal}, 22(8): 7971--7978.

\bibitem[{Rashed-Al-Mahfuz et~al.(2021)Rashed-Al-Mahfuz, Moni, Uddin, Alyami,
  Summers, and Eapen}]{rashed2021deep}
Rashed-Al-Mahfuz, M.; Moni, M.~A.; Uddin, S.; Alyami, S.~A.; Summers, M.~A.;
  and Eapen, V. 2021.
\newblock A deep convolutional neural network method to detect seizures and
  characteristic frequencies using epileptic electroencephalogram (EEG) data.
\newblock \emph{IEEE Journal of Translational Engineering in Health and
  Medicine}, 9: 1--12.

\bibitem[{Saab et~al.(2020)Saab, Dunnmon, R{\'e}, Rubin, and
  Lee-Messer}]{saab2020weak}
Saab, K.; Dunnmon, J.; R{\'e}, C.; Rubin, D.; and Lee-Messer, C. 2020.
\newblock Weak supervision as an efficient approach for automated seizure
  detection in electroencephalography.
\newblock \emph{NPJ digital medicine}, 3(1): 1--12.

\bibitem[{Saboksayr, Mateos, and Cetin(2021)}]{saboksayr2021eeg}
Saboksayr, S.~S.; Mateos, G.; and Cetin, M. 2021.
\newblock EEG-based emotion classification using graph signal processing.
\newblock In \emph{ICASSP 2021-2021 IEEE International Conference on Acoustics,
  Speech and Signal Processing (ICASSP)}, 1065--1069. IEEE.

\bibitem[{Saichand et~al.(2021)}]{saichand2021epileptic}
Saichand, N.~V.; et~al. 2021.
\newblock Epileptic seizure detection using novel Multilayer LSTM Discriminant
  Network and dynamic mode Koopman decomposition.
\newblock \emph{Biomedical Signal Processing and Control}, 68: 102723.

\bibitem[{Shah et~al.(2018)Shah, Von~Weltin, Lopez, McHugh, Veloso,
  Golmohammadi, Obeid, and Picone}]{shah2018temple}
Shah, V.; Von~Weltin, E.; Lopez, S.; McHugh, J.~R.; Veloso, L.; Golmohammadi,
  M.; Obeid, I.; and Picone, J. 2018.
\newblock The temple university hospital seizure detection corpus.
\newblock \emph{Frontiers in neuroinformatics}, 12: 83.

\bibitem[{Sharma, Rai, and Tewari(2018)}]{sharma2018epileptic}
Sharma, A.; Rai, J.; and Tewari, R. 2018.
\newblock Epileptic seizure anticipation and localisation of epileptogenic
  region using EEG signals.
\newblock \emph{Journal of Medical Engineering \& Technology}, 42(3): 203--216.

\bibitem[{Shen et~al.(2022)Shen, Wen, Song, and Li}]{shen2022eeg}
Shen, M.; Wen, P.; Song, B.; and Li, Y. 2022.
\newblock An EEG based real-time epilepsy seizure detection approach using
  discrete wavelet transform and machine learning methods.
\newblock \emph{Biomedical Signal Processing and Control}, 77: 103820.

\bibitem[{Shoeibi et~al.(2021)Shoeibi, Khodatars, Ghassemi, Jafari, Moridian,
  Alizadehsani, Panahiazar, Khozeimeh, Zare, Hosseini-Nejad
  et~al.}]{shoeibi2021epileptic}
Shoeibi, A.; Khodatars, M.; Ghassemi, N.; Jafari, M.; Moridian, P.;
  Alizadehsani, R.; Panahiazar, M.; Khozeimeh, F.; Zare, A.; Hosseini-Nejad,
  H.; et~al. 2021.
\newblock Epileptic seizures detection using deep learning techniques: a
  review.
\newblock \emph{International Journal of Environmental Research and Public
  Health}, 18(11): 5780.

\bibitem[{Surges et~al.(2021)Surges, Shmuely, Dietze, Ryvlin, and
  Thijs}]{surges2021identifying}
Surges, R.; Shmuely, S.; Dietze, C.; Ryvlin, P.; and Thijs, R.~D. 2021.
\newblock Identifying patients with epilepsy at high risk of cardiac death:
  signs, risk factors and initial management of high risk of cardiac death.
\newblock \emph{Epileptic Disorders}, 23(1): 17--39.

\bibitem[{Tang et~al.(2021)Tang, Dunnmon, Saab, Zhang, Huang, Dubost, Rubin,
  and Lee-Messer}]{tang2021self}
Tang, S.; Dunnmon, J.; Saab, K.~K.; Zhang, X.; Huang, Q.; Dubost, F.; Rubin,
  D.; and Lee-Messer, C. 2021.
\newblock Self-Supervised Graph Neural Networks for Improved
  Electroencephalographic Seizure Analysis.
\newblock In \emph{International Conference on Learning Representations}.

\bibitem[{Thuwajit et~al.(2021)Thuwajit, Rangpong, Sawangjai, Autthasan,
  Chaisaen, Banluesombatkul, Boonchit, Tatsaringkansakul, Sudhawiyangkul, and
  Wilaiprasitporn}]{thuwajit2021eegwavenet}
Thuwajit, P.; Rangpong, P.; Sawangjai, P.; Autthasan, P.; Chaisaen, R.;
  Banluesombatkul, N.; Boonchit, P.; Tatsaringkansakul, N.; Sudhawiyangkul, T.;
  and Wilaiprasitporn, T. 2021.
\newblock EEGWaveNet: Multiscale CNN-Based Spatiotemporal Feature Extraction
  for EEG Seizure Detection.
\newblock \emph{IEEE Transactions on Industrial Informatics}, 18(8):
  5547--5557.

\bibitem[{Tong, Faloutsos, and Pan(2006)}]{tong2006fast}
Tong, H.; Faloutsos, C.; and Pan, J.-Y. 2006.
\newblock Fast random walk with restart and its applications.
\newblock In \emph{Sixth international conference on data mining (ICDM'06)},
  613--622. IEEE.

\bibitem[{Wagh and Varatharajah(2020)}]{wagh2020eeg}
Wagh, N.; and Varatharajah, Y. 2020.
\newblock Eeg-gcnn: Augmenting electroencephalogram-based neurological disease
  diagnosis using a domain-guided graph convolutional neural network.
\newblock In \emph{Machine Learning for Health}, 367--378. PMLR.

\bibitem[{Wang et~al.(2020)Wang, Liang, Wang, Zhang, He, Ma, Ruan, Wu, Hong,
  and Shen}]{wang2020weighted}
Wang, J.; Liang, S.; Wang, Y.; Zhang, Y.; He, D.; Ma, J.; Ruan, C.; Wu, Y.;
  Hong, X.; and Shen, J. 2020.
\newblock A weighted overlook graph representation of eeg data for absence
  epilepsy detection.
\newblock In \emph{2020 IEEE International Conference on Data Mining (ICDM)},
  581--590. IEEE.

\bibitem[{Xie et~al.(2022)Xie, Xu, Zhang, Wang, and Ji}]{xie2022self}
Xie, Y.; Xu, Z.; Zhang, J.; Wang, Z.; and Ji, S. 2022.
\newblock Self-supervised learning of graph neural networks: A unified review.
\newblock \emph{IEEE Transactions on Pattern Analysis and Machine
  Intelligence}.

\bibitem[{Yan et~al.(2018)Yan, Dong, Mu, Liu, Chen, Deng, Wang, and
  Zhao}]{yan2018positive}
Yan, T.; Dong, X.; Mu, N.; Liu, T.; Chen, D.; Deng, L.; Wang, C.; and Zhao, L.
  2018.
\newblock Positive classification advantage: tracing the time course based on
  brain oscillation.
\newblock \emph{Frontiers in human neuroscience}, 11: 659.

\bibitem[{Zeng and Xie(2021)}]{zeng2021contrastive}
Zeng, J.; and Xie, P. 2021.
\newblock Contrastive self-supervised learning for graph classification.
\newblock In \emph{Proceedings of the AAAI Conference on Artificial
  Intelligence}, 10824--10832.

\bibitem[{Zhang et~al.(2021)Zhang, Lu, Zhan, and Zhang}]{zhang2021semi}
Zhang, H.; Lu, G.; Zhan, M.; and Zhang, B. 2021.
\newblock Semi-supervised classification of graph convolutional networks with
  Laplacian rank constraints.
\newblock \emph{Neural Processing Letters}, 1--12.

\bibitem[{Zheng et~al.(2021)Zheng, Jin, Liu, Chi, Phan, and
  Chen}]{zheng2021generative}
Zheng, Y.; Jin, M.; Liu, Y.; Chi, L.; Phan, K.~T.; and Chen, Y.-P.~P. 2021.
\newblock Generative and contrastive self-supervised learning for graph anomaly
  detection.
\newblock \emph{IEEE Transactions on Knowledge and Data Engineering}.

\end{thebibliography}

\end{document}